\documentclass[hidelinks,hypertexnames=false,sn-mathphys,Numbered,oneside]{sn-jnl}

\usepackage[utf8]{inputenc}

\usepackage{newtxtext}
\usepackage{manyfoot}
\usepackage{xurl}           
\usepackage{graphicx}%
\usepackage{multirow}%
\usepackage{amsmath,amssymb,amsfonts}%
\usepackage{newtxmath}
\usepackage[title]{appendix}%
\usepackage{xcolor}%
\usepackage[scr=rsfso]{mathalfa}
\usepackage{textcomp}%
\usepackage{booktabs}%
\usepackage{algorithm}%
\usepackage{algorithmicx}%
\usepackage{algpseudocode}%
\usepackage{listings}%
\usepackage{setspace,array}
\newcolumntype{P}[1]{>{\raggedright\arraybackslash\setstretch{0.9}}p{#1}} 
\lstdefinestyle{prompt}{%
  basicstyle=\ttfamily\small,
  breaklines=true,
  columns=fullflexible,
  frame=single,
  frameround=tttt,
  rulecolor=\color{black!20},
  backgroundcolor=\color{black!2},
  numbers=none
}
\lstdefinestyle{jsonprompt}{%
  style=prompt,
  language=json,
  showstringspaces=false
}
\usepackage{subcaption}
\usepackage{caption} 
\usepackage{balance}  
\usepackage{array,booktabs}
\usepackage{siunitx}
\usepackage{tabularx}
\usepackage{upgreek}
\usepackage{tikz}
\usepackage{graphicx}
\usetikzlibrary{angles, quotes, shapes.geometric, positioning, backgrounds}
\newcolumntype{L}{>{\raggedright\arraybackslash}X}   
\begin{document}

\title{Recov-Vision: Linking Street View Imagery and Vision-Language Models for Post-Disaster Recovery}


\author*[1]{\fnm{Yiming} \sur{Xiao}}\email{yxiao@tamu.edu}
\author[1]{\fnm{Archit} \sur{Gupta}}\email{archit.gupta@tamu.edu}
\author[1]{\fnm{Miguel} \sur{Esparza}}\email{mte1224@tamu.edu}
\author[1]{\fnm{Yu-Hsuan} \sur{Ho}}\email{yuhsuanho@tamu.edu}
\author[2]{\fnm{Antonia} \sur{Sebastian}}\email{asebastian@unc.edu} 
\author[2]{\fnm{Hannah} \sur{Weas}}\email{hweas@unc.edu}
\author[2]{\fnm{Rose} \sur{Houck}}\email{rosehouck@unc.edu}
\author[1]{\fnm{Ali} \sur{Mostafavi}}\email{amostafavi@civil.tamu.edu}
\affil*[1]{\orgdiv{UrbanResilience.AI Lab, Zachry Department of Civil and Environmental Engineering}, \orgname{Texas A\&M University}, \orgaddress{\street{199 Spence St}, \city{College Station}, \state{Texas}, \postcode{77840}, \country{USA}}}

\affil[2]{\orgdiv{Department of Earth, Marine and Environmental Sciences}, \orgname{The University of North Carolina at Chapel Hill}, \orgaddress{\street{104 South Rd}, \city{Chapel Hill}, \state{North Carolina}, \postcode{27514}, \country{USA}}}

\abstract{Building-level occupancy after disasters is critical for triage, inspections, utility re-energization, and equitable resource allocation. Overhead imagery provides rapid coverage but often misses facade and access cues that determine habitability, while street-view imagery captures those details but is sparse and hard to align with parcels. We present Recov-Vision, a street-level, language-guided framework that links panoramic video to parcels, rectifies views to facades, and elicits interpretable attributes (e.g., entry blockage, temporary coverings, localized debris) that drive two decision strategies: a transparent one-stage rule and a two-stage design separating perception from conservative reasoning. Evaluated across two post-Hurricane Helene surveys, the reasoning approach attains higher recall with comparable overall agreement to a one-stage baseline and reproduces the ground-truth net recovery (+14), while intermediate attributes and spatial diagnostics expose where and why residual errors occur. The pipeline provides auditable, scalable occupancy assessments suitable for integration into geospatial and emergency-management workflows.}


\maketitle

\section{Introduction}\label{sec:intro}

\subsection{Motivation \& problem statement}\label{sec:motivation}
Rapid, building-level occupancy status after a disaster is a time-critical input for triage, utility restoration, inspections, temporary shelter planning, and equitable resource allocation. Overhead (nadir) imagery arrives quickly and covers large areas, but its roof-dominant perspective often overlooks the very cues that determine whether people can safely inhabit a building---open or blocked entries, temporary coverings, localized debris, and ad-hoc repairs. Even state-of-the-art post-disaster overhead methods struggle when habitability hinges on facade and access conditions rather than roof damage alone \cite{shen_bdanet_2022, kaur_largescale_2023, rahnemoonfar_floodnet_2021}. 

Street-based visual surveys and door-to-door assessments supply those missing signals, yet they are inherently sparse and viewpoint-dependent. They are constrained to the road network and acquisition schedules, and their alignment with parcel inventories is often noisy \cite{yang_hyperlocal_2025, xue_post-hurricane_2024, cinnamon_panoramic_2021, martell_open-source_2024}. As a result, analysts face a slow, manual reconciliation of overhead and ground evidence that is difficult to reproduce or audit at scale. This process typically occurs only once immediately after impact, rather than being repeated to track recovery over time.

Recent progress in vision-language models (VLMs) could provide a mechanism for turning heterogeneous visual evidence into operational decisions without hand-crafted taxonomies for every new event. By combining open-vocabulary recognition, instruction following, and cross-modal reasoning, VLMs can connect small visual cues with higher-level statements about building condition and use \cite{weng_vision-language_2025}. While early examples establish feasibility---ranging from change reasoning \cite{deng_changechat_2025} to language-conditioned elevation estimation \cite{ho_integrated_2025}---they stop short of a reproducible, street-level pipeline centered on occupancy inference. 

A key limitation of past work is the inability to stitch together complementary evidence across views to turn ambiguous signals into explicit intermediate attributes. Small indicators such as vehicle mix, tarps behind foliage, or shadows around doorways need to be promoted into structured, inspectable attributes so that an assessor can verify how the model moved from evidence to judgment \cite{liang_openfacades_2025}. Furthermore, because domain-tuned aerial and street-view methods already achieve high accuracy, claims of progress require statistical comparisons, spatial localization of errors, and analysis of failure cases tied to viewpoint and occlusion. These gaps become more pressing as agencies seek scalable, auditable assessments to defend resource-allocation decisions \cite{paulik_local_2025}.

\subsection{Research objectives and significance}\label{sec:objectives}
This work develops a street-level, language-guided framework for post-disaster building occupancy assessment. We combine panoramic imaging, automated facade rectification, and vision-language reasoning to produce parcel-level occupancy labels and recovery trajectories. 

The primary objective is to create a reproducible pipeline that transforms raw street-level video into auditable occupancy decisions. Unlike "black box" end-to-end classifiers, our approach emphasizes intermediate evidence (facade rectification, attribute extraction) and conservative reasoning to monitor parcel-level change across repeat visits.

The significance of this study lies in two areas: measurement fidelity and operational targeting. By employing a two-stage reasoning strategy, the pipeline improves recall and accurately reproduces ground-truth net recovery rates, whereas standard baselines tend to overstate recovery. This yields credible metrics that decision-makers can act on. Spatial diagnostics reveal significant clustering in occupancy states and residual errors. By generating accuracy maps, the framework allows agencies to pinpoint pockets for targeted quality assurance and quality control (QA/QC) and field tasking rather than relying on diffuse, random reviews.

\subsection{Contributions}\label{sec:contributions}
To address the gaps in cross-view integration, attribute interpretability, and rigorous validation, this paper makes four main contributions:

\begin{enumerate}
    \item A reproducible street-level capture-to-label pipeline. We document an end-to-end workflow that links 5.6K panoramic videos to parcels, estimates heading from GPS windows, and uses a simple yaw formulation to rectify frames into facade-centered views. These geometry and processing steps are fully specified to enable replication and auditing.
    
    \item Interpretable, policy-tunable occupancy inference. We compare two decision designs using a shared vision extractor: a transparent one-stage rule and a two-stage perception-to-reasoning approach. By separating perception from logic, thresholds can be tuned for conservatism without retraining the vision component.
    
    \item Temporally and spatially aware evaluation. We introduce a change framework (Recovered, Deteriorated, Stable) to quantify net recovery across campaigns. We utilize global Moran's \textit{I} to diagnose the geographic concentration of errors, enabling spatially targeted quality control.
    
    \item Statistically grounded validation. We report paired bootstrap confidence intervals and McNemar tests to rigorously compare strategies. Our error taxonomy shows that the two-stage approach reduces ambiguity, guiding reviewers to a small, well-defined subset of difficult cases.
\end{enumerate}

\subsection{Paper organization}\label{sec:organization}
Section~\ref{sec:related} reviews vision-language foundations and existing damage assessment pipelines. Section~\ref{sec:method} details the data collection, frame-parcel linkage, and decision strategies. Section~\ref{sec:results} reports classification performance, recovery metrics, and spatial error patterns. Section~\ref{sec:discussion} discusses deployment implications and limitations, followed by conclusions in Section~\ref{sec:conclusion}.

\section{Related work}\label{sec:related}

\subsection{Aerial and satellite imagery for post-disaster damage}
Deep learning with overhead imagery has become the standard for rapid, wide-area damage mapping. Synthesizing this progress, \citet{wang_deep_2024} survey architectures and open challenges, noting a shift from simple classification to semantic segmentation. Early pipelines successfully used convolutional models on open aerial data to map impacts \citep{gupta_deep_2021}, yet operational analyses have consistently documented how nadir views are confounded by environmental factors—clouds, canopy, and shadows—that obscure key cues and complicate severity estimation \citep{qing_operational_2022}. Furthermore, the lack of large, balanced datasets remains a persistent obstacle \citep{wang_building_2022}. While curated benchmarks such as FloodNet \citep{rahnemoonfar_floodnet_2021} have standardized protocols, their scale is limited. To address the gap between raw detection and decision-making, typology-driven formulations have recently argued for actionable categories aligned with civil recovery needs rather than coarse severity scores \citep{xiao_damagecat_2025}.

Recent modeling advances have focused on improving representation capacity and temporal reasoning. Hierarchical transformers have emerged as a solution for scaling to large areas while preserving fine-grained detail \citep{kaur_largescale_2023}. In the domain of change detection, Siamixformer introduced a fully transformer-based Siamese architecture for bi-temporal reasoning \citep{mohammadian_siamixformer_2023}, while BDANet combined multiscale features with cross-directional attention to better capture local and global context \citep{shen_bdanet_2022}. Beyond single architectures, ensemble learning has been explored to enhance robustness across diverse hydrologic and geographic terrains \citep{roohi_developing_2024}. 

Research has also expanded into end-to-end mapping systems that integrate imagery with GIS layers \citep{braik_automated_2024} and urban remote-sensing workflows that pair deep learning with post-classification correction \citep{techapinyawat_integrated_2024}. Recognizing the limits of satellite resolution, UAV and oblique-aerial studies have developed tailored models to quantify damage from lower-altitude perspectives \citep{calantropio_deep_2021, gong_deep_2021, khankeshizadeh_novel_2024}. Some efforts have even coupled UAV data with GIS-based information systems for facade assessment \citep{chen_gis-based_2023}.

Despite these advances, overhead approaches remain fundamentally limited by their viewing angle. They struggle when habitability hinges on facade and access cues—such as boarded windows or blocked doors—rather than roof integrity \citep{ho_flood-damagesense_2025}, necessitating complementary ground-level evidence.

\subsection{Street-view imagery for urban condition and post-event assessment}
Street-view imagery (SVI) provides the high-resolution, facade-level detail required for occupancy analysis. \citet{cinnamon_panoramic_2021} review the methodological considerations of roadside acquisition, while open-source pipelines have been introduced to streamline the ingest, filtering, and sampling of SVI for research \citep{martell_open-source_2024}. Outside of disaster contexts, SVI has powered city-scale analyses of facade attributes \citep{zhong_city-scale_2021}, pavement defect mapping \citep{maniat_deep_2021, kong_automatic_2022, wang_automatic_2021}, and the identification of social vulnerability indicators \citep{ogawa_deep_2023}.

In disaster response, bi-temporal SVI has been applied to hyperlocal change assessment \citep{yang_hyperlocal_2025}, and multi-modal fusion studies have shown that combining SVI with structured attributes improves damage prediction accuracy \citep{xue_post-hurricane_2024}. To address occlusion and viewing angle challenges, recent work has leveraged multiple SVI vantage points to improve robustness \citep{gu_multiview_2025}, or fused SVI with remote sensing for broader flood risk modeling \citep{chen_multi-modal_2022, xing_flood_2023}. Operational guidance on reconnaissance-oriented SVI collection—covering route design and coverage—has also helped bridge the gap between research and deployment \citep{errett_street_nodate}.

While SVI captures the necessary visual data, most existing methods rely on "black box" classifiers or simple late fusion. They lack the structured, transparent reasoning required to defend resource allocation decisions, particularly when views are occluded or oblique.

\subsection{Vision-language and multimodal foundations}
Vision-language models (VLMs) offer a mechanism to process visual evidence using open-vocabulary recognition and cross-modal reasoning. \citet{weng_vision-language_2025} survey emerging applications in remote sensing, noting the potential for VLMs to capture context-specific features like accessibility and structural condition. \citet{deng_changechat_2025} demonstrated explanation-driven change reasoning on paired images, while \citet{wang_disasterm3_2025} introduced a multi-hazard, multi-sensor benchmark to evaluate these models under diverse conditions.

Recent studies have begun to tailor VLMs to specific built-environment tasks. \citet{ho_integrated_2025} used language-conditioned estimation to determine lowest-floor elevation from street views, and \citet{liang_openfacades_2025} utilized large models to enrich facade attribute datasets at scale. In retrieval tasks, cross-view geo-localization has been used to surface relevant ground photos for overhead scenes \citep{sogi_disaster_2024}. Beyond floods, multimodal systems have supported wildfire triage and infrastructure description \citep{esparza_automated_2025, zhang_vision-enhanced_2025}. Complementary efforts on knowledge fusion aim to mitigate data scarcity by combining imagery with expert priors \citep{wei_effective_nodate}.

These studies establish the feasibility of VLM-based analysis but generally stop at detection or description. There remains a lack of integrated pipelines that use VLMs to drive a rigorous, reproducible, and auditable occupancy inference system capable of tracking recovery trajectories over time.

\section{Methodology}\label{sec:method}

This study proposes a pipeline that combines Street View Imagery (SVI) with Vision Language Models (VLMs) to infer building occupancy status in post-disaster scenarios. The method integrates geospatial data collection and processing, computer vision techniques, and natural language reasoning to assess whether residential buildings are occupied following extreme disaster events. The approach consists of four key components: (1) collection and preprocessing of panoramic street-view video data with GPS tracking to generate building-specific facade views, (2) development of a vision-language model prompting strategies that extract visual damage indicators and infer occupancy status, (3) temporal inference modeling to track occupancy changes over time, and (4) comprehensive evaluation against ground truth data. This multi-modal framework enables scalable, objective assessment of post-disaster building conditions using street-level imagery and state-of-the-art language models.

\subsection{Data collection}\label{sec:3.1}

As a test case, we collected imagery from two field visits to areas impacted by Hurricane Helene. Helene made landfall as a Category 4 hurricane in the Big Bend region of Florida on September 26, 2024, before traveling inland and stalling over the North Carolina/Tennessee border on September 27--28. In total, Helene is estimated to have generated more than \$78 billion in economic losses \citep{re2024natural} and at least 250 fatalities. Of these, 95 were directly attributable to freshwater flooding (which includes landslides and debris flows) in North Carolina, South Carolina, and Tennessee, making it the deadliest hurricane to make landfall in the continental United States since Hurricane Katrina (2005) \citep{papin_climatology_2017}. Notably, Helene's heavy precipitation (i.e., 20--30 inches, corresponding to a 1000-year rainfall event in western North Carolina) over a region characterized by steep topography generated widespread flooding, damaging tens of thousands of homes and cutting off entire communities from access to critical facilities such as water, sewer, electricity, telecommunications, and healthcare \citep{re_natural_2024}. The southern Appalachian Mountains exhibit particularly low flood insurance penetration (i.e., only 1.5\% of buildings had coverage \citep{amorim_flooding_2025}), raising questions about how quickly the region will recover and what household recovery will look like.

\begin{figure}[!htbp]
    \centering
    \includegraphics[width=0.9\textwidth]{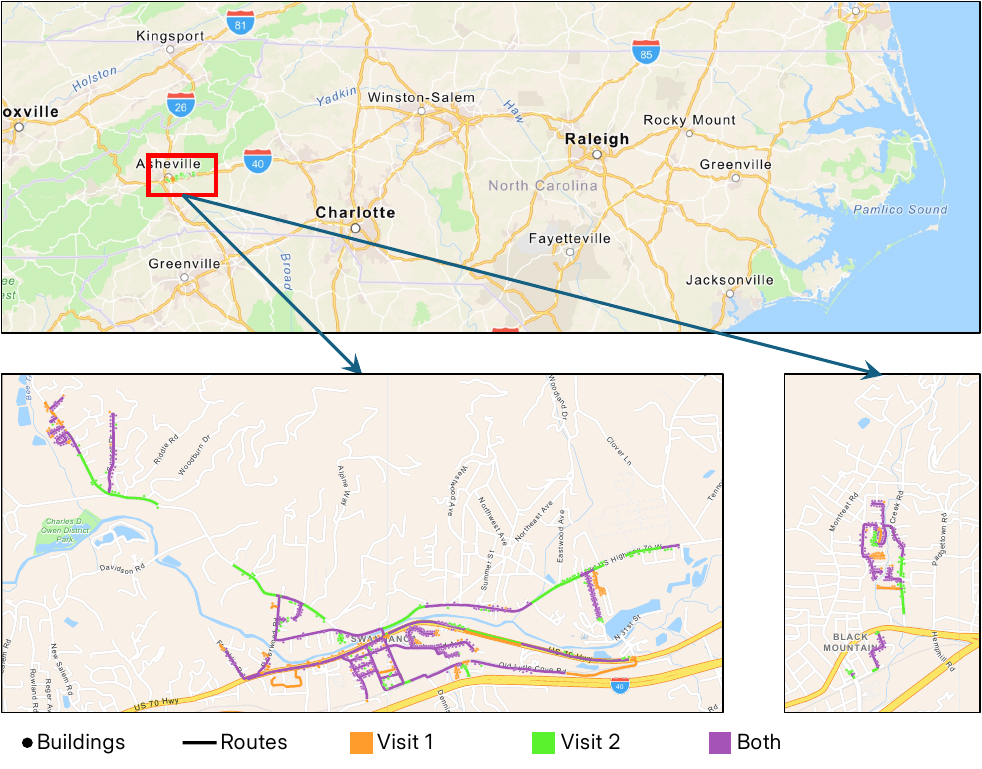}
    \caption{Study area in broader Asheville area, North Carolina, showing residential parcels and drive routes used for panoramic data collection. Dots indicate buildings and lines indicate drive routes. Orange indicates routes/buildings from Visit 1, Green Visit 2, and Purple from both.}
    \label{fig:study-area}
\end{figure}

Our study focused on several flooded neighborhoods in Swannanoa and Black Mountain in Buncombe County, North Carolina. (Swannanoa is a census-designated place located between the City of Asheville and the Town of Black Mountain.) Both communities are located along the Swannanoa River, which is a tributary of the French Broad River. The key survey areas are shown in Figure~\ref{fig:study-area}. Prior post-disaster surveys (e.g., Red Cross assessments and county emergency management reports \citep{Esri_COVID_Dashboard}) identified these neighborhoods as among the most heavily affected communities in the region. Our field reconnaissance in December 2024 confirmed the high density of visible damage, making them appropriate testbeds for property-level monitoring. The damaged areas comprised a mix of suburban and semi-rural parcels and had relatively accessible road networks, enabling repeated street-level imaging to support temporal analysis of recovery. In addition to the initial reconnaissance visit conducted in December 2024, we collected street-level data across two field campaigns in March 2025 and June 2025.

During each field visit, we collected street‐level, 360° imagery using a GoPro Max panoramic camera mounted on a rigid roof-rack mast (approximately 2\,m above ground level) on a passenger vehicle. Videos were recorded at the camera’s native panoramic resolution (5.6K) at 30\,fps to support frame‐level alignment with GPS. Each recording was accompanied by a synchronized per‐video GPS log (CSV) capturing latitude/longitude, speed, and course, enabling subsequent geospatial matching of frames to reference parcels (Section~\ref{sec:3.2}). The camera was leveled and oriented with the vehicle’s leftward heading to better capture the building facades, with horizon lock enabled to stabilize the panoramic stream. For each campaign, operators maintained a simple time‐stamped field log noting start/stop points, detours, obstructions, and environmental conditions to facilitate post hoc quality control and alignment. All data were captured from public rights‐of‐way; no entry onto private property occurred. To protect privacy, personally identifying details (e.g., faces and license plates) were blurred or removed prior to the dissemination of imagery.

\subsection{Data pre-processing}\label{sec:3.2}

This approach assembles a street-level video and GPS dataset to generate parcel-linked facade views suitable for downstream vision-language prompting and temporal inference. The workflow includes matching reference buildings to vehicle GPS tracks at frame resolution, extracting the corresponding video frames, estimating vehicle heading, and dewarping panoramic images to centered rectilinear views.

Reference points are converted to a projected CRS for metric operations (EPSG:5070) and buffered by 25 m to form search regions. For each GPS file, latitude/longitude samples are transformed to the same CRS and converted to point geometries. An R-tree spatial index over buffered reference regions retrieves candidate buildings per GPS point; among candidates, the nearest reference by Euclidean distance is kept. For each match, we record the reference index, the video identifier, the vehicle coordinates from the GPS sample, and a frame index defined as the row position in the GPS file (frame number). Matched rows are merged back to the reference attributes (reprojected to EPSG:4326), duplicates are removed, a unique ObjectId (1..\(n\)) is assigned, and the video name is cleaned to a base identifier. The result is saved as a table for frame extraction.

\subsubsection{View rectification}

\begin{figure}[!htbp]
    \centering
    \begin{subfigure}[b]{0.48\textwidth}
        \centering
        \includegraphics[width=\textwidth]{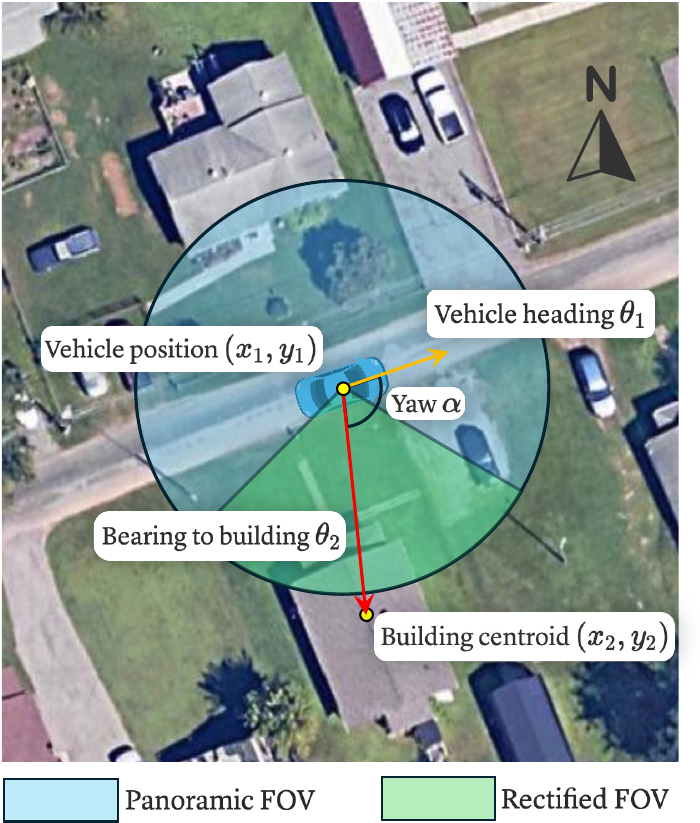}
        \caption{Geometric relationship for view rectification.}
        \label{fig:rectification-a}
    \end{subfigure}
    \hfill
    \begin{subfigure}[b]{0.48\textwidth}
        \centering
        \includegraphics[width=\textwidth]{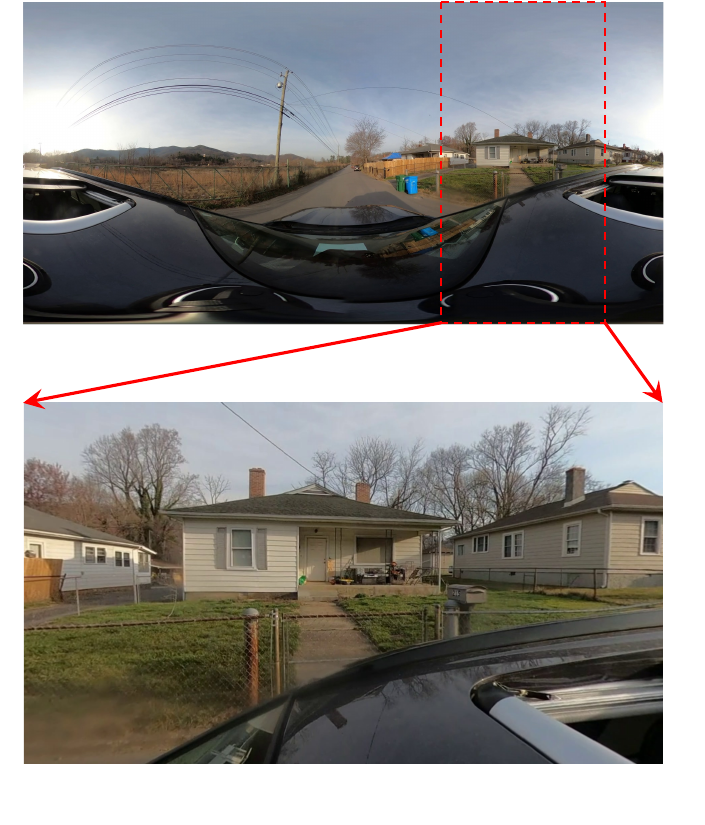}
        \caption{Rectification of a panoramic image to a planar projection.}
        \label{fig:rectification-b}
    \end{subfigure}
    \caption{Visualization of the view rectification process. (a) The top-down view illustrates the geometric parameters used for rectification, including the vehicle's position \((x_1, y_1)\) and heading \((\vec{\theta}_1)\), and the building's centroid \((x_2, y_2)\). The bearing to the building \((\vec{\theta}_2)\) is calculated, and the yaw angle \(\alpha\) orients the rectified field of view (green) toward the building. (b) A raw 360-degree panoramic image is processed to generate a rectified, planar view of the building facade, which is then used for analysis.}
    \label{fig:rectification}
\end{figure}

Frame extraction groups videos by frame number, finding the corresponding media file by trying common extensions and name variants, and obtaining duration and frame rate. Requested frame indices are validated against the total frame count; invalid requests are dropped with reasons recorded. Each valid frame number is converted to a timestamp (frame number/fps) and extracted with FFmpeg as a single high-quality JPEG\@ Frames are named by ObjectId and written to a frames directory.

Since the vehicle's heading information is unavailable, for each matched record, the process reopens the corresponding GPS track and approximates the vehicle heading at the target frame using nearby samples: averaging a small window of points before and after the frame (for example, \(\pm\) 15 frames, approximately 0.5 s at 30 fps), then computes the bearing from the mean ``before'' position to the mean ``after'' position. This bearing (degrees) is stored as the vehicle's current heading. The enriched table is saved for subsequent dewarping.

Using building and vehicle coordinates, the algorithm computes the bearing from the camera to the building and derives a yaw to center the facade: \(\text{yaw } \alpha = (\text{bearing to building } \theta_{2}) - (\text{vehicle heading } \theta_{1})\) plus a camera-specific calibration offset (\(\theta_{\text{calib}}\), \(-90^{\circ}\) in our runs). Yaw is normalized to \([-180,180]\) degrees. For panoramic frames (wide aspect ratio indicating equirectangular input), the system dewarps to a rectilinear view using FFmpeg's v360 filter with fixed parameters (horizontal field of view = 90\(^{\circ}\), aspect ratio = 16:9, output width = 1920 px). Non-panoramic frames are passed through unchanged. Outputs are written to a separate directory for rectified frames, and existing outputs may be skipped to allow incremental runs. Figure~\ref{fig:rectification} illustrates the geometric relationships and rectification process.

Quality control tracks invalid frame indices, missing GPS rows, and dewarping skips, and records whether a frame was rectified or copied. The process yields a raw frame and a rectified frame centered on the target facade, together with a table containing ObjectId, reference attributes, video name, frame number, vehicle coordinates, and estimated heading per matched building-view.

The data processing pipeline includes several key calculations. The bearing calculation between two geographic points \((lat_1, lon_1)\) and \((lat_2, lon_2)\) follows the standard great-circle bearing formula:
\begin{align}
\Delta lon &= lon_2 - lon_1 \\
y &= \sin(\Delta lon) \cdot \cos(lat_2) \\
x &= \cos(lat_1) \cdot \sin(lat_2) - \sin(lat_1) \cdot \cos(lat_2) \cdot \cos(\Delta lon) \\
\theta &= \text{atan2}(y, x) \cdot \frac{180}{\uppi}
\end{align}
where \(\theta\) is normalized to \([0, 360]\) degrees. For spatial matching, reference buildings are buffered by 25\,m in projected coordinates (EPSG:5070), and an R-tree spatial index enables efficient retrieval of candidate matches within buffer bounds. Vehicle heading estimation averages GPS positions within a temporal window of \(\pm 15\) frames (\(\approx 0.5\)s at 30\,fps) and applies the bearing formula to the mean positions. Frame number \(f\) is converted to timestamp \(t\) via \(t = f / \text{fps}\), where fps is the video frame rate. For view rectification, the yaw angle \(\alpha\) that centers the building in the rectified view is computed as:
\begin{equation}
\alpha = (\theta_{2} - \theta_{1} + \theta_{\text{calib}}) \bmod 360
\end{equation}
where \(\theta_{2}\) is the bearing from camera to building, \(\theta_{1}\) is the vehicle heading, and \(\theta_{\text{calib}} = -90^{\circ}\) is the camera calibration offset. The yaw is normalized to \([-180, 180]\) degrees before application to the FFmpeg v360 filter.

The resulting rectified facade views constitute the inputs to the VLM prompting module (Section\,\ref{sec:3.3}) and the temporal comparison and recovery mapping (Section\,\ref{sec:3.4}). Calibration offset, field-of-view, and window size are tunable; reported analyses use values set a priori for the study area and collection period.

To evaluate the performance of the method, rectified facade images were independently labeled by two annotators using a three-class system (Occupied, Not Occupied, Uncertain) guided by a written code book defining observable cues (e.g., debris piles, access obstructions, emergency markings, active repairs, vehicle presence). The GoPro operator reviewed all of the labels against field logs and raw panoramas to verify context and visibility and made the final determination of the occupancy status for each frame. Frames labeled as Uncertain or with insufficient visibility were excluded from primary accuracy metrics. For locations visited in both campaigns, labels were assigned per campaign independently to support temporal comparisons.

\subsection{VLM prompting strategy}\label{sec:3.3}

\begin{figure}[!htbp]
    \centering
    \includegraphics[width=0.7\textwidth]{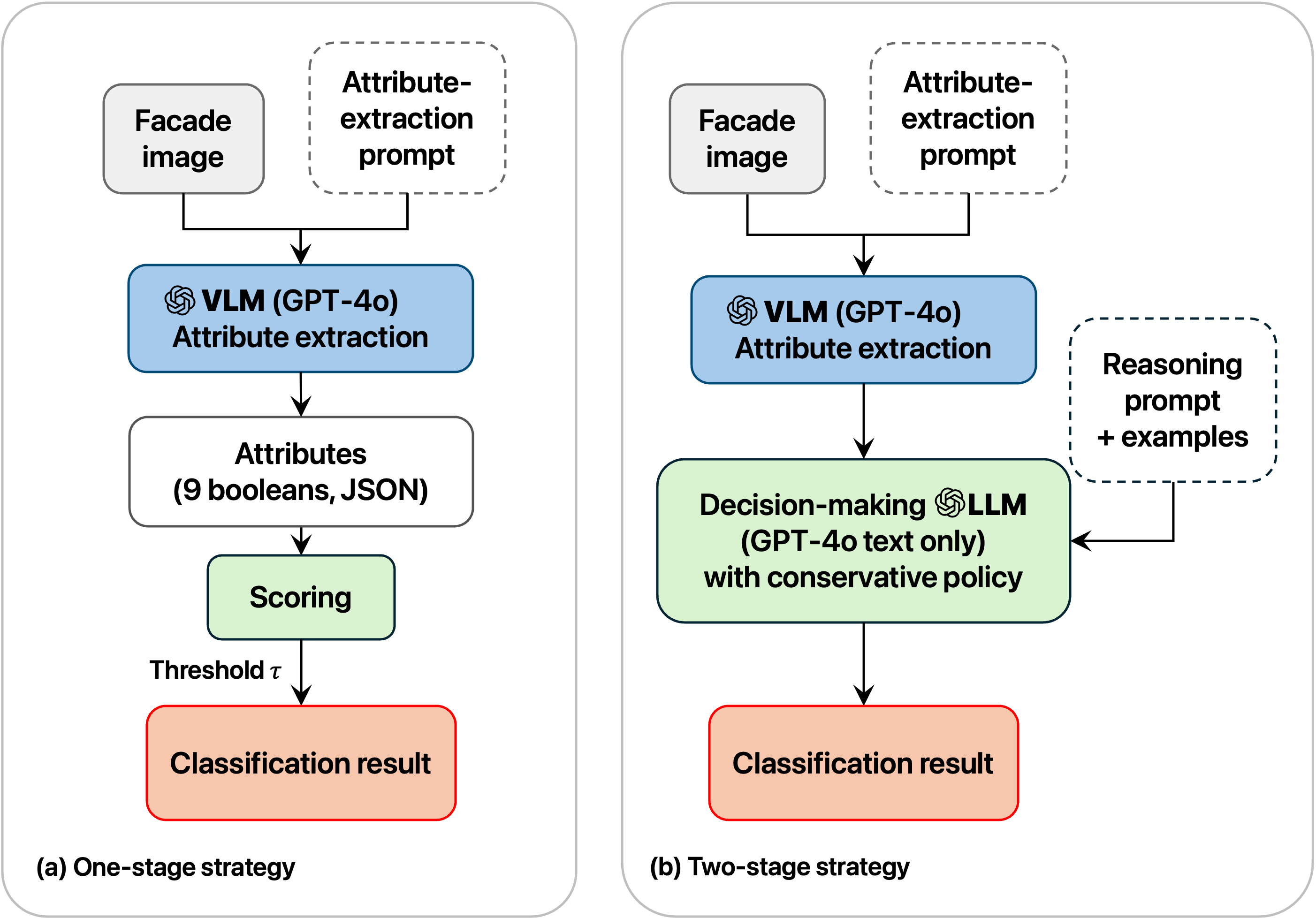}
    \caption{Overview of the two prompting strategies for building occupancy classification. (a) Single-stage baseline: a vision-language model (VLM) extracts nine visual attributes, and a deterministic scoring rule with threshold \(\tau\) produces the final label. (b) Two-stage strategy: the same VLM extracts the attributes; a text-only reasoning LLM applies explicit rules and few-shot exemplars to generate a conservative final decision.}
    \label{fig:vlm-workflow}
\end{figure}

The proposed building occupancy assessment employs vision-language models to analyze the rectified facade images from the pre-processing pipeline. Two distinct strategies are implemented and compared: a single-stage vision-only approach and a two-stage vision-reasoning pipeline (Figure~\ref{fig:vlm-workflow}).

The first strategy, which serves as a baseline, applies a fixed-count rule over eight risk indicators (all schema fields except \texttt{vehicle\_presence}). Let $r$ denote the number of risk indicators that are \texttt{true}, and let $v{=}1$ if a curbside vehicle is visible (\texttt{vehicle\_presence=true}) and $0$ otherwise. The decision policy implemented in code is equivalent to
\begin{equation}
\text{predict Not Occupied}\iff r - v \ge \tau,\quad \text{with }\tau=2.
\end{equation}

Thus, the effective threshold is two when no vehicle is visible and three when a vehicle is present (i.e., one point is subtracted for visible vehicles to reflect the prior that cars correlate with occupancy). Table~\ref{tab:baseline-threshold} summarizes the rules used in all experiments.

\begin{table}[!htbp]
\captionsetup{justification=raggedright,singlelinecheck=false}
\centering
\caption{Baseline threshold ($\tau$) with vehicle adjustment. Risk indicators are: \texttt{house\_destruction}, \texttt{structural\_damage}, \texttt{exterior\_debris}, \texttt{open\_doors\_windows}, \texttt{site\_accessible=false}, \texttt{exterior\_mud}, \texttt{emergency\_markings}, \texttt{major\_repairs}.}
\label{tab:baseline-threshold}
\normalsize
\begin{tabular}{>{\raggedright\arraybackslash}p{5cm} >{\raggedright\arraybackslash}p{3cm} >{\raggedright\arraybackslash}p{5.2cm}}
\toprule
\textbf{Vehicle visible?} & \textbf{Effective $\tau$ (risk indicators needed)} & \textbf{Decision summary} \\
\midrule
No (\texttt{vehicle\_presence=false}) & $\tau=2$ & Not Occupied if $r\ge2$; else Occupied \\
Yes (\texttt{vehicle\_presence=true}) & $\tau=3$ & Not Occupied if $r\ge3$; else Occupied \\
\bottomrule
\end{tabular}
\end{table}

The choice of $\tau{=}2$ reflects a conservative rule: isolated single cues (e.g., a small debris pile) are insufficient for a Not Occupied decision without corroboration, and visible vehicles raise the required count by one to reduce false positives in occupied but repairing homes. This keeps the baseline transparent and tunable; future work will sweep the threshold to empirically validate the operating point.

The second strategy introduces a separate decision-making stage. In the feature-extraction stage, the same multi-modal model analyzes each rectified image with a prompt that requests only observable visual evidence (the nine attributes above), without inferring occupancy. In the subsequent decision stage, we use the language model (GPT-4o) in text-only mode to map the extracted attributes to a final occupancy label via explicit reasoning rules and few-shot learning examples (see Appendix~\ref{sec:appendix}, Listing~\ref{lst:decision-fewshot}). These rules and exemplars (nine cases covering diverse damage and occupancy scenarios) encode domain expertise. The decision policy is intentionally conservative, favoring ``Not Occupied'' when evidence is ambiguous or suggests potential habitability concerns. This two-stage approach is referred to as the ``Two-stage'' strategy in the following analysis.

\begin{table}[!htbp]
\centering
\caption{Attribute schema used in VLM extraction: definitions and typical observable cues.}
\label{tab:attribute-defs}
\normalsize
\renewcommand{\arraystretch}{1.15} 
\begin{tabular}{P{3.4cm} P{11.8cm}}
\toprule
\textbf{Attribute in prompts} & \textbf{Typical visual cues (examples)} \\
\midrule
\texttt{house\_destruction} & Severe or total loss of habitability: collapse, missing major wall sections, roof largely gone, house off foundation. \\
\texttt{structural\_damage} & Visible structural or envelope damage short of destruction: partial roof damage, compromised porch, broken/boarded windows, leaning walls, major siding loss. \\
\texttt{exterior\_debris} & Disaster or cleanup debris in yard/drive/curb: piles of damaged contents, construction rubble, spoiled furniture, scattered materials (exclude small ordinary clutter). \\
\texttt{open\_doors\_windows} & Door or window visibly open (ajar) or missing, suggesting exposure; excludes normal closed / reflective windows where interior not visible. \\
\texttt{site\_accessible} & Parcel frontage unobstructed (TRUE) vs blocked (FALSE) by fencing, gate, caution tape, barricade, large dumpster/trailer preventing normal entry. \\
\texttt{exterior\_mud} & Distinct flood or sediment deposits: mud lines on siding, dried silt on walls/steps/yard, uniform brown coating inconsistent with normal dirt. \\
\texttt{emergency\_markings} & Official post-disaster markings: spray-painted search/inspection tags (e.g., X-codes), placards, colored notices on doors/windows; exclude ordinary numbers or graffiti. \\
\texttt{major\_repairs} & Active or recent substantial repair work: tarps on roof, exposed framing, scaffolding, large construction materials (lumber stacks, shingles bundles), contractor equipment. \\
\texttt{vehicle\_presence} & One or more vehicles visibly parked in driveway, curb, or immediate lot area (cars, trucks, vans). Construction-only equipment without a personal vehicle may be excluded. \\
\bottomrule
\end{tabular}
\renewcommand{\arraystretch}{1.0} 
\end{table}

This two-stage approach offers several methodological advantages over direct classification. The separation of perception and reasoning enables systematic evaluation of visual feature extraction accuracy independent of decision logic. The structured attribute representation provides interpretable intermediate outputs that facilitate error analysis and model debugging. The explicit reasoning stage allows incorporation of domain knowledge and conservative decision policies without requiring retraining of the vision components. Additionally, the standardized attribute schema enables consistent evaluation across different building types and damage patterns. The attributes used in the VLM extraction stage and their typical visual cues are summarized in Table~\ref{tab:attribute-defs}. Note that Emergency responders often spray-paint a large “X” on building exteriors during search-and-rescue operations, indicating that the structure was inspected and often that it was damaged or uninhabitable.

Both strategies are evaluated on the same dataset of rectified building facade images to assess their relative performance in post-disaster occupancy assessment tasks.

All language-model calls used the OpenAI API (chat/responses endpoint), which accepts role-based messages and images. For fairness, both strategies shared the same vision attribute-extraction stage (identical multimodal model, schema, and prompt; see Appendix~\ref{sec:appendix}, Listings~\ref{lst:vision-only-prompt}--\ref{lst:vision-only-json}); only the decision stage differed. The vision extractor used a system message and a user message (image plus schema) with \texttt{temperature}=0 and \texttt{max\_tokens}=200. The decision stage used a text-only call with \texttt{temperature}=0 and \texttt{max\_tokens}=5, along with the system message ``\emph{You are a helpful assistant that analyzes images and outputs structured JSON.}'' and the few-shot exemplars in Listing~\ref{lst:decision-fewshot}. No explicit \texttt{response\_format} (JSON mode) was requested; instead, JSON structure was enforced by instruction. The one-stage baseline then applied a deterministic scoring rule to the extracted attributes.

No random seed was set, and other sampling parameters used default values (e.g., \texttt{top\_p}). The code does not implement automatic JSON validation or retry-on-failure; model responses were consumed as returned.

\subsection{Temporal comparison and recovery mapping}\label{sec:3.4}

This stage derives parcel-level occupancy change descriptors between the two field campaigns for each inference strategy and prepares data structures for recovery mapping. Inputs are per-parcel visit labels from the VLM prompting stage (Section~\ref{sec:3.3}) for both the single-stage (vision scoring) and two-stage (vision and reasoning) strategies, together with human ground-truth annotations for the subset of parcels manually labeled in each campaign. After visibility and quality filtering, parcel identifiers are joined across visits in chronological order (V1 for Visit 1 and V2 for Visit 2). When multiple rectified frames exist for a parcel within a visit, each strategy first produces frame-level labels; a visit-level label is then assigned by majority vote. Ties, mutually contradictory evidence, or any frame flagged Uncertain produce a conservative Not Occupied assignment; if all usable frames are Uncertain, the visit label is set to Uncertain and excluded from primary accuracy and change metrics. The same consolidation rule is applied to ground-truth frame annotations.

For parcels with valid (Occupied or Not Occupied) labels at both visits, we define a change class given by the ordered pair (V1, V2). The four resulting classes are: Recovered (from Not Occupied to Occupied), Deteriorated (from Occupied to Not Occupied), Stable--Occupied (stay as Occupied), and Stable--Not Occupied (stay as Not Occupied), summarized in Table~\ref{tab:transitions}. Parcels with an Uncertain or missing label at either visit are excluded from change summaries for that strategy but still counted in coverage statistics (fraction of parcels yielding a usable change).
\begin{table}[!htbp]
\centering
\caption{Parcel occupancy change classes between Visit 1 (V1) and Visit 2 (V2).}
\label{tab:transitions}
\normalsize
\begin{tabular*}{0.6\textwidth}{@{\hspace{0.9em}}l@{\extracolsep{\fill}}l@{\hspace{0.9em}}}
\toprule
\textbf{Change} & \textbf{Definition (V1 to V2)} \\
\midrule
Recovered & Not Occupied to Occupied \\
Deteriorated & Occupied to Not Occupied \\
Stable--Occupied & Occupied to Occupied \\
Stable--Not Occupied & Not Occupied to Not Occupied \\
\bottomrule
\end{tabular*}
\end{table}

Structured outputs of this stage are: (1) a visit-level label table containing, for each parcel and strategy (and ground truth where available), the consolidated V1 and V2 labels with exclusion flags; and (2) a parcel change table enumerating change classes for parcels with valid labels at both visits. These outputs serve as inputs to the evaluation metrics and numerical results in Section~\ref{sec:results}.

\subsection{Model Performance}

We report 95\% confidence intervals (CIs) for Precision, Recall, F1, and Cohen's \(\kappa\) via paired bootstrap resampling at the parcel level. We emphasize Recall and F1 because the dataset is moderately imbalanced between Occupied and Not Occupied, making these metrics more informative than raw accuracy. Cohen’s \(\kappa\) complements them by accounting for agreement beyond chance, which is important in imbalanced settings. Between-method differences are defined as $\Delta=\text{two-stage}-\text{one-stage}$ and are assessed with paired bootstrap tests (reporting $\Delta$ and 95\% CIs) and McNemar's test applied separately to Visit~1 and Visit~2 contingency tables. We interpret $p$-values as evidence for or against a difference in error rates, while effect sizes and their CIs guide practical conclusions.

\section{Results}\label{sec:results}

\subsection{VLM prompting pipeline performance}

\begin{figure}[!htbp]
    \centering
    \includegraphics[width=\textwidth]{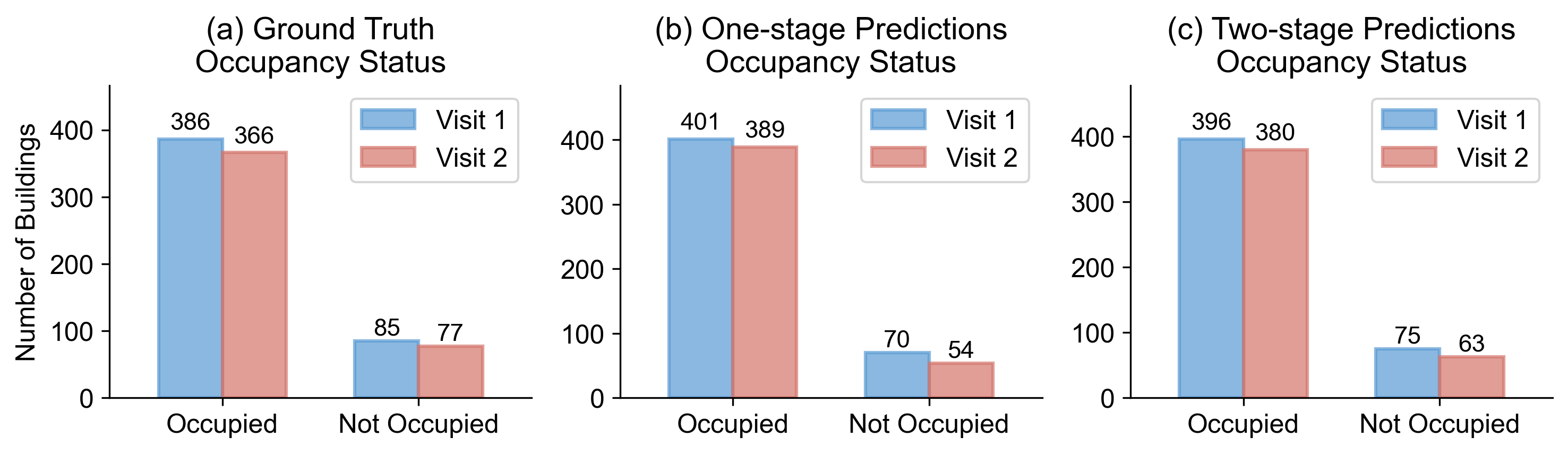}
    \caption{Occupancy status overview for Ground Truth, Two-stage predictions, and One-stage predictions across both visits.}
    \label{fig:occupancy-overview}
\end{figure}

The original dataset included 578 building facades from Visit 1 and 621 from Visit 2. After filtering out images labeled as ``Unknown'' due to obstruction, blur, poor viewing angle, or no building present, the usable sample comprised 471 facades for Visit 1 and 443 for Visit 2. These filtered counts define the set of parcels used for model evaluation across both strategies.
Model performance is compared to ground truth in several ways. The Ground Truth, predictions by the Two-stage strategy, and by the One-stage strategy are shown in Figure~\ref{fig:occupancy-overview}. The figure shows the occupancy status for each parcel at both visits. Ground truth vacancies declined from 85 to 77 ($-8$; $-9.4\%$), indicating a modest decrease in vacancies between the two field surveys. The Two-stage strategy shows the same trend but with greater magnitude (75 to 63; -12; $-16.0\%$), while the One-stage strategy shows an even larger drop (70 to 54; $-16$; $-22.9\%$). However, both models overestimate occupancy relative to ground truth at each visit (Two-stage: $-10$ V1 / $-14$ V2; One-stage: $-15$ V1 / $-23$ V2), with larger overestimates at V2 consistent with under-predicted vacancy. Total counted parcels differ between visits because parcels with Uncertain or missing labels are excluded, but the relative shifts still show an occupancy contraction overstated by the One-stage baseline. These distributional biases signal the net recovery bias and change agreement results reported below.

\begin{figure}[!htbp]
    \centering
    \includegraphics[width=0.9\textwidth]{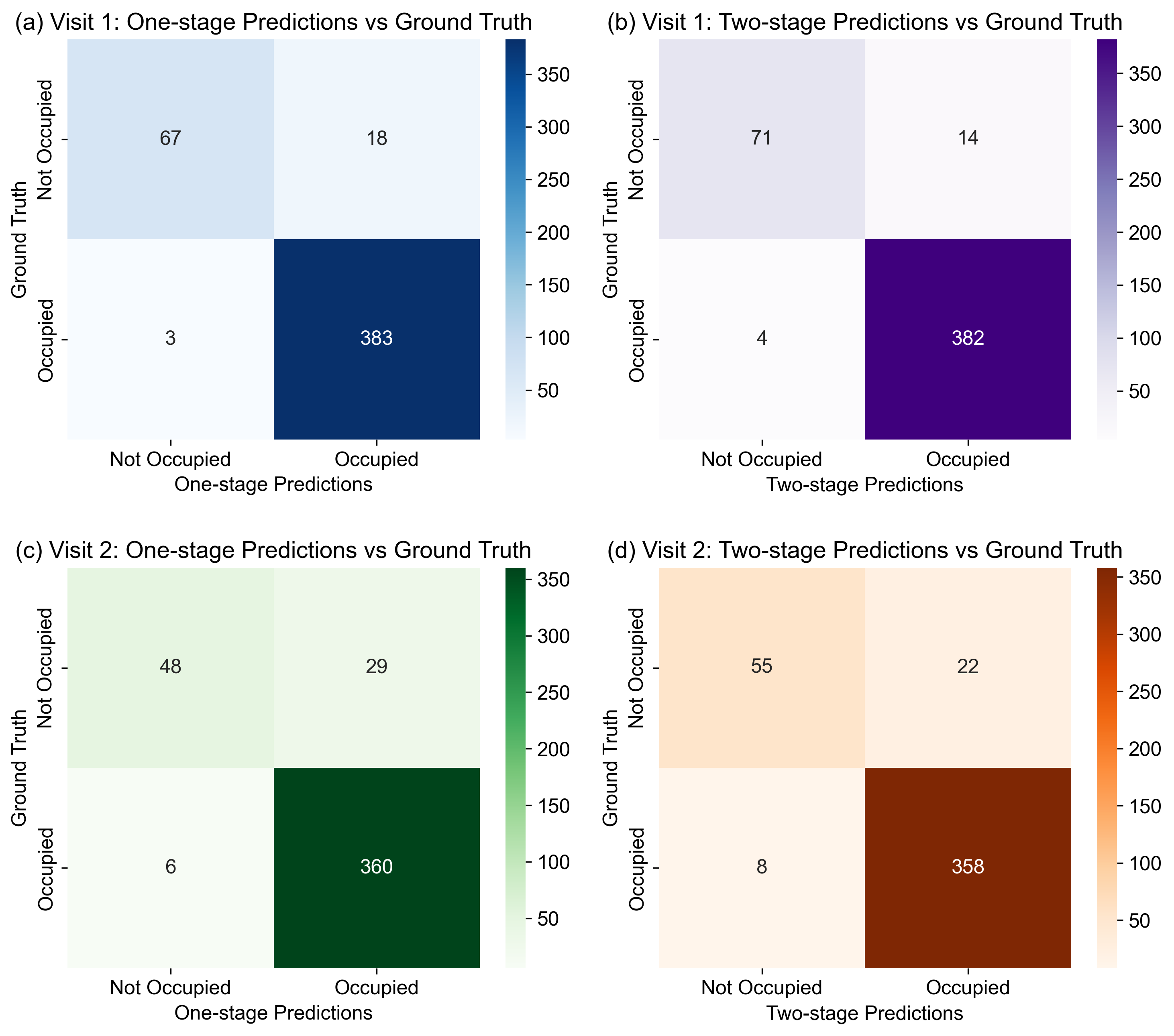} 
    \caption{Confusion matrices for both visits, comparing model predictions to ground truth. Not Occupied is treated as a positive class.}
    \label{fig:confusion-matrices}
\end{figure}

For each strategy, we compute per-visit confusion matrices (shown in Figure~\ref{fig:confusion-matrices}; Not Occupied is treated as positive class) and report accuracy, precision, recall, F1, and Cohen's $\kappa$ in Table~\ref{tab:method-metrics-ci}. Table~\ref{tab:method-metrics-ci} summarizes performance with 95\% CIs, and Table~\ref{tab:stats-tests} reports paired significance tests.

\begin{table}[!htbp]
\caption{Statistical significance tests for method comparisons. McNemar tests compare per-visit error patterns; paired bootstrap tests compare metric differences.}
\label{tab:stats-tests}
\normalsize
\begin{tabular}{lccr}
\toprule
\textbf{Test} & \textbf{Statistic} & \textbf{p-value} & \textbf{Interpretation} \\
\midrule
McNemar (Visit 1) & 0.571 & 0.450 & Not significant \\
McNemar (Visit 2) & 0.000 & 1.000 & Not significant \\
Paired Bootstrap ($\Delta F_{1}$) & +0.026 & 0.188 & Not significant \\
Paired Bootstrap ($\Delta \kappa$) & +0.029 & 0.226 & Not significant \\
\bottomrule
\end{tabular}
\end{table}

\begin{table}[!htbp]
\caption{Method performance comparison with bootstrap confidence intervals. Not Occupied is treated as the positive class.}
\label{tab:method-metrics-ci}
\normalsize
\begin{tabular}{lccccr}
\toprule
\textbf{Metric} & \textbf{One-stage} & \textbf{One-stage 95\% CI} & \textbf{Two-stage} & \textbf{Two-stage 95\% CI} & \textbf{Difference} \\
\midrule
Precision & 0.943 & [0.886, 1.000] & 0.927 & [0.873, 0.998] & -0.016 \\
Recall & 0.728 & [0.648, 0.862] & 0.781 & [0.707, 0.902] & +0.053 \\
F1-Score & 0.822 & [0.759, 0.900] & 0.848 & [0.795, 0.930] & +0.026 \\
Cohen's $\kappa$ & 0.789 & [0.718, 0.884] & 0.818 & [0.756, 0.916] & +0.029 \\
\bottomrule
\end{tabular}
\end{table}

As shown in Table~\ref{tab:method-metrics-ci}, the two-stage strategy exhibits a modest precision-recall trade-off relative to the one-stage baseline. It achieves higher recall at the cost of slightly lower precision, resulting in numerically higher F1 and Cohen's $\kappa$ scores. However, as shown in Table~\ref{tab:stats-tests}, these differences are not statistically significant. Both McNemar's tests on per-visit error patterns and paired bootstrap tests on the performance metrics yield non-significant p-values, with confidence intervals for the differences spanning zero. This suggests that while the two-stage method yields slightly higher recall and agreement, the two strategies are statistically indistinguishable at the current operating point.

\begin{figure}[!htbp]
    \centering
    \includegraphics[width=0.8\textwidth]{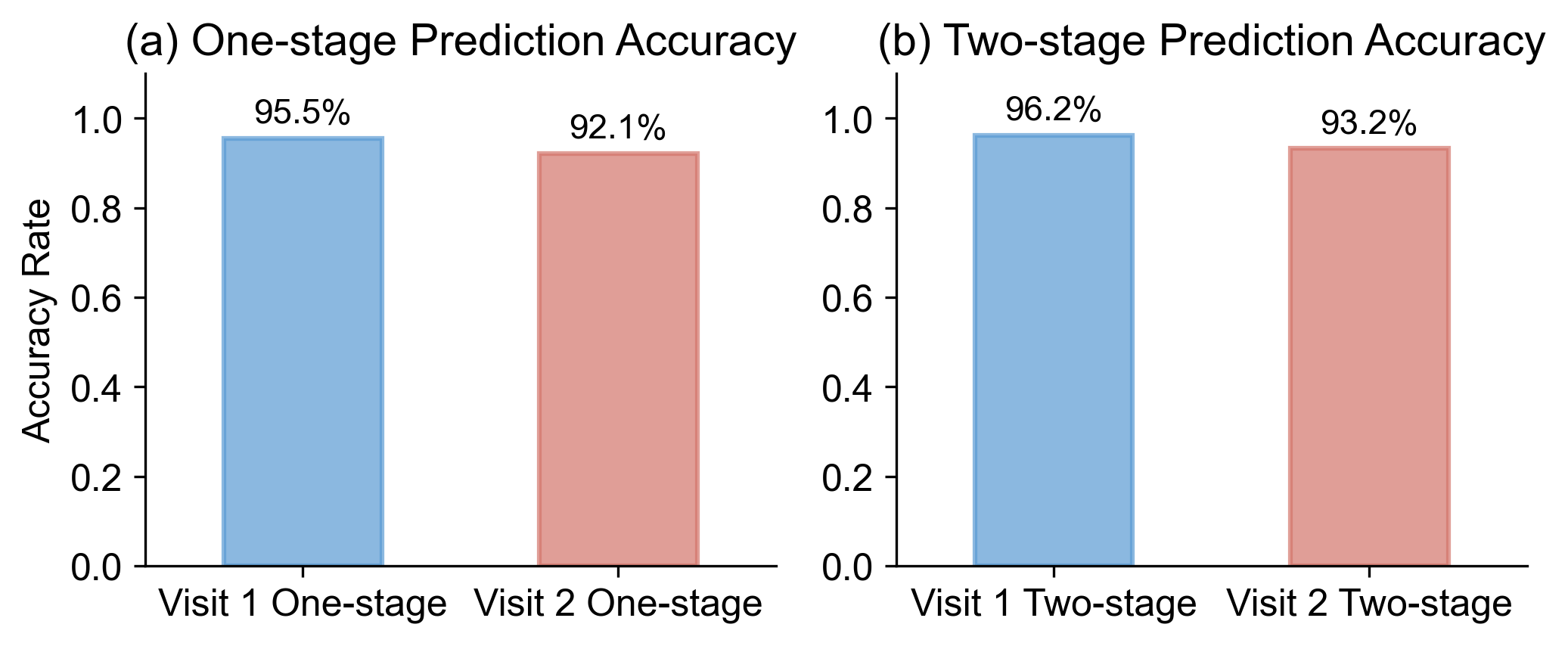}
    \caption{Per-visit overall accuracy: (a) One-stage strategy (95.5\% V1, 92.1\% V2); (b) Two-stage strategy (96.2\% V1, 93.2\% V2).}
    \label{fig:occupancy-accuracy}
\end{figure}

Overall accuracies (Figure~\ref{fig:occupancy-accuracy}) decline modestly from V1 to V2 for both strategies (Two-stage -3.0 pp (percentage point) vs One-stage -3.4 pp), with the Two-stage maintaining a consistent ~0.7--1.1 pp advantage each visit. The parallel drop suggests a shared increase in classification difficulty (e.g., more subtle post-repair cues) rather than strategy-specific drift.

Direct comparison of the two strategies focuses on expected advantages of the two-stage (vision and reasoning) method: fewer false Not Occupied labels at a single visit (improved Occupied recall), higher precision for Deteriorated changes by propagating fewer spurious vacancies, improved precision for Recovered (fewer artificial recoveries caused by an incorrect Not Occupied baseline at V1), and higher overall change $\kappa$. Disagreements between strategies are flagged for manual review and spatially highlighted.

\begin{figure}[!htbp]
    \centering
    \begin{subfigure}[t]{0.48\textwidth}
        \centering
        \includegraphics[width=\textwidth]{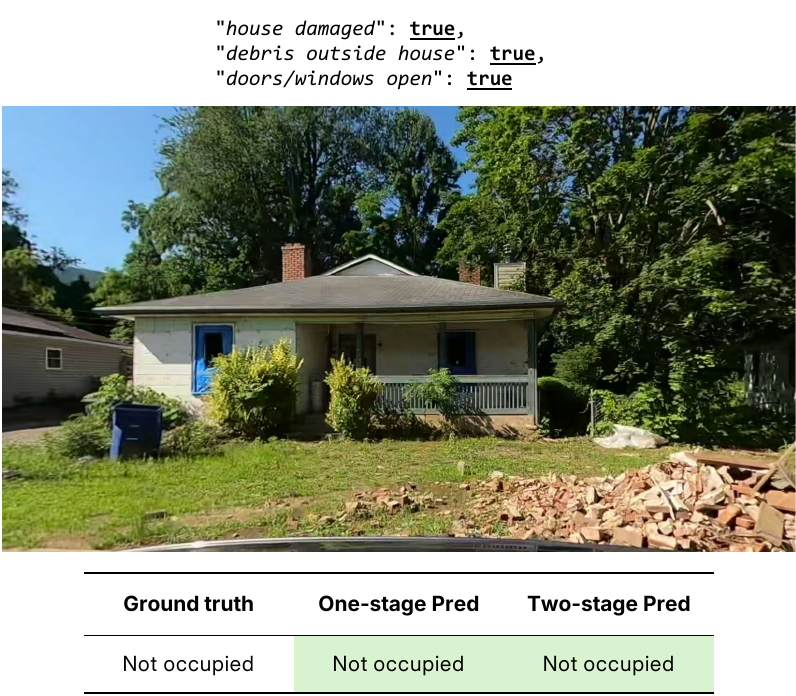}
        \caption{Case 1: both strategies correct (Not occupied).}
        \label{fig:example1}
    \end{subfigure}
    \hfill
    \begin{subfigure}[t]{0.48\textwidth}
        \centering
        \includegraphics[width=\textwidth]{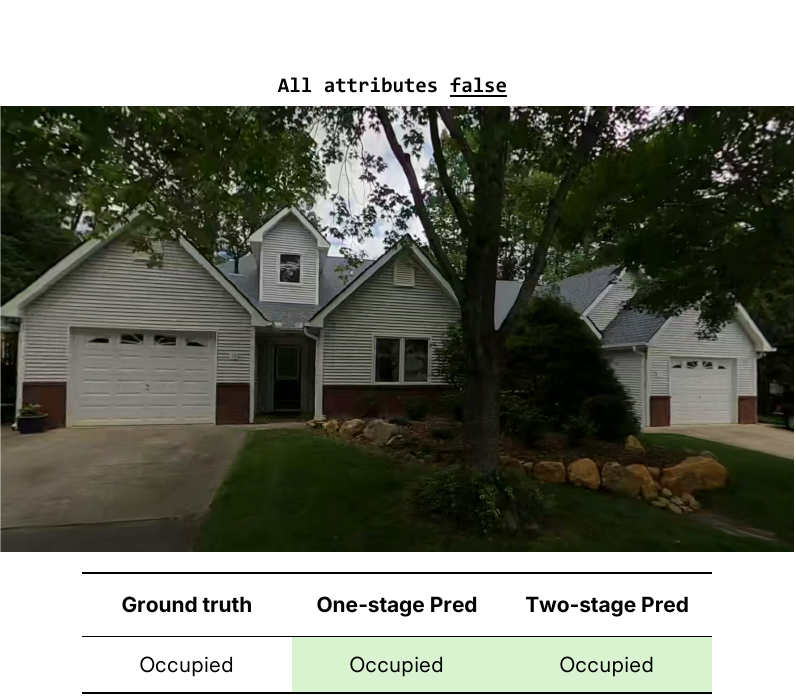}
        \caption{Case 2: both strategies correct (Occupied).}
        \label{fig:example2}
    \end{subfigure}

    \begin{subfigure}[t]{0.48\textwidth}
        \centering
        \includegraphics[width=\textwidth]{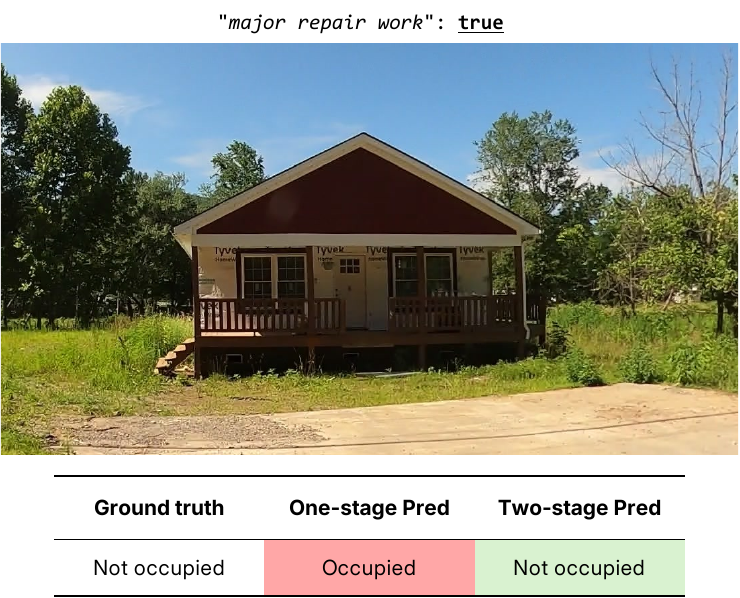}
        \caption{Case 3: disagreement between strategies.}
        \label{fig:example3}
    \end{subfigure}
    \hfill
    \begin{subfigure}[t]{0.48\textwidth}
        \centering
        \includegraphics[width=\textwidth]{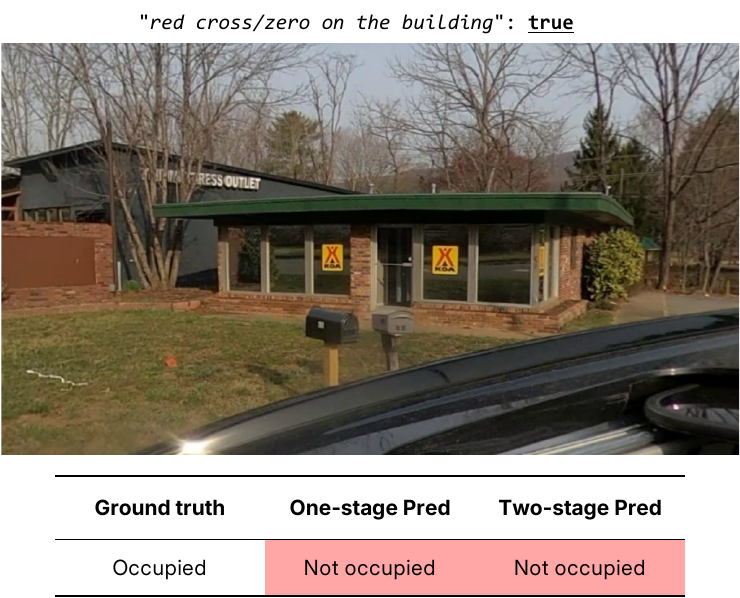}
        \caption{Case 4: both strategies misclassified.}
        \label{fig:example4}
    \end{subfigure}

    \caption{Examples of rectified building facades with ground truth and model predictions. Sub-figures illustrate a range of outcomes: (a--b) correct predictions, (c) strategy disagreement, and (d) residual error, which is probably caused by mistakenly recognizing the logo on the building as ``X'' marks. Only attributes that are true are shown above each figure.}
    \label{fig:qual-examples}
\end{figure}

\begin{figure}[!htbp]
    \centering
    \includegraphics[width=0.6\textwidth]{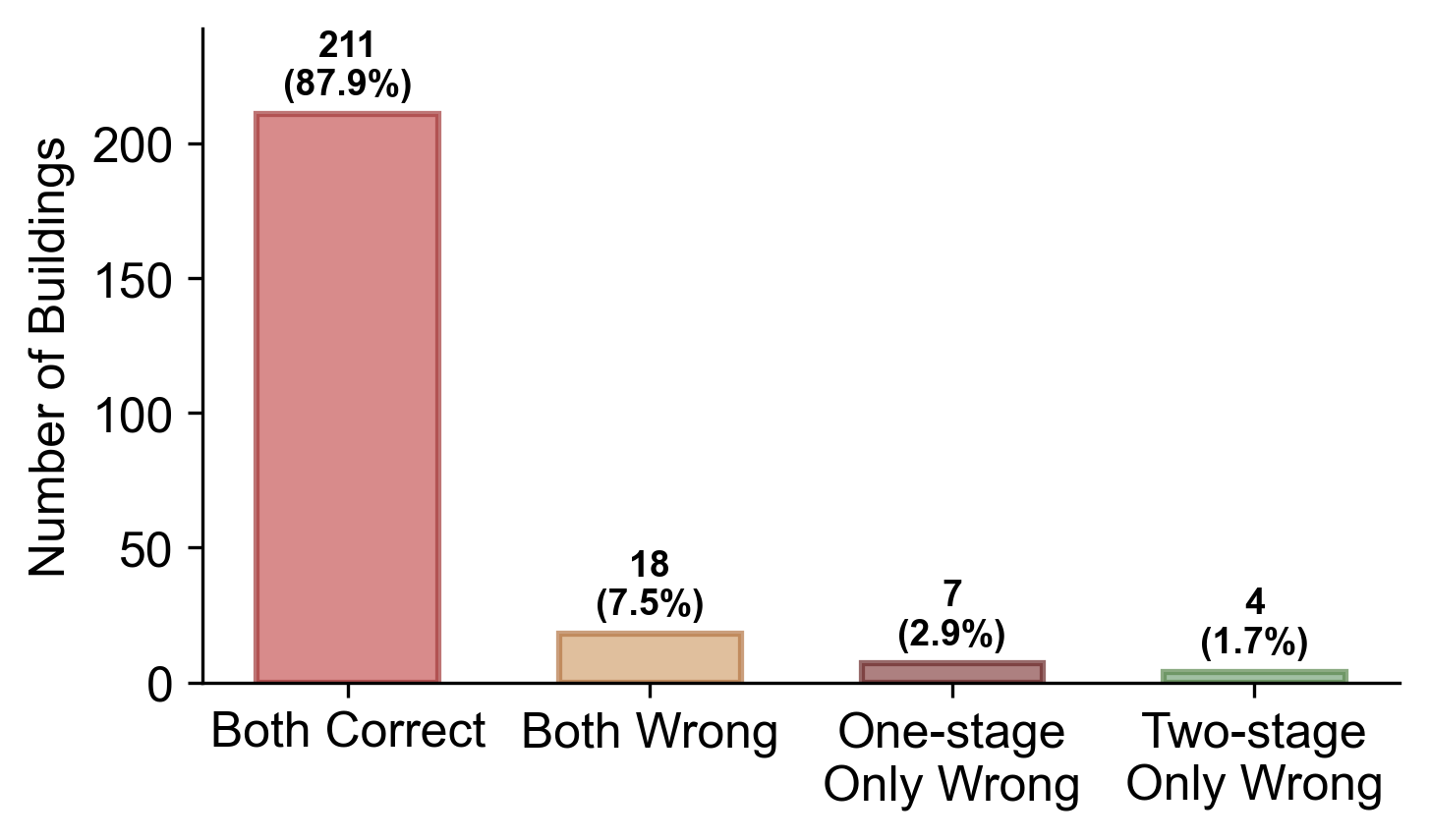}
    \caption{Error pattern analysis comparing correctness across parcels: Both Correct, Both Wrong, only One-stage wrong, or only Two-stage wrong (counts with percentages).}
    \label{fig:error-patterns}
\end{figure}

Figure \ref{fig:qual-examples} shows four representative cases. In panel (a), both strategies correctly classify the parcel as Not Occupied, supported by clear exterior debris and boarded openings. In panel (b), both strategies correctly label an intact home as Occupied. Panel (c) illustrates a disagreement: the one-stage baseline predicts Occupied while the two-stage reasoning strategy assigns Not Occupied, reflecting its more conservative decision rule. Finally, panel (d) shows a residual error where both strategies misclassify due to confounding visual cues (in this case, the model mistakenly interpreted the logo on the door as "X" markers). Together, these examples highlight how the methods align with ground truth in most straightforward cases, but also where conservative reasoning can alter outcomes or where both models still struggle.

Figure~\ref{fig:error-patterns} shows that most parcels (211; 87.9\%) are correctly classified by both strategies. Joint errors are more frequent (18; 7.5\%), and asymmetric errors slightly favor the two-stage strategy (one-stage only wrong: 7; 2.9\% vs two-stage only wrong: 4; 1.7\%). This pattern corroborates the two-stage model's higher recall and agreement, while highlighting that residual challenges concentrate in a small set of ambiguous parcels (Both Wrong) suitable for targeted prompt refinement or data augmentation.

\subsection{Occupancy change analysis}

\begin{figure}[!htbp]
    \centering
    \includegraphics[width=0.8\textwidth]{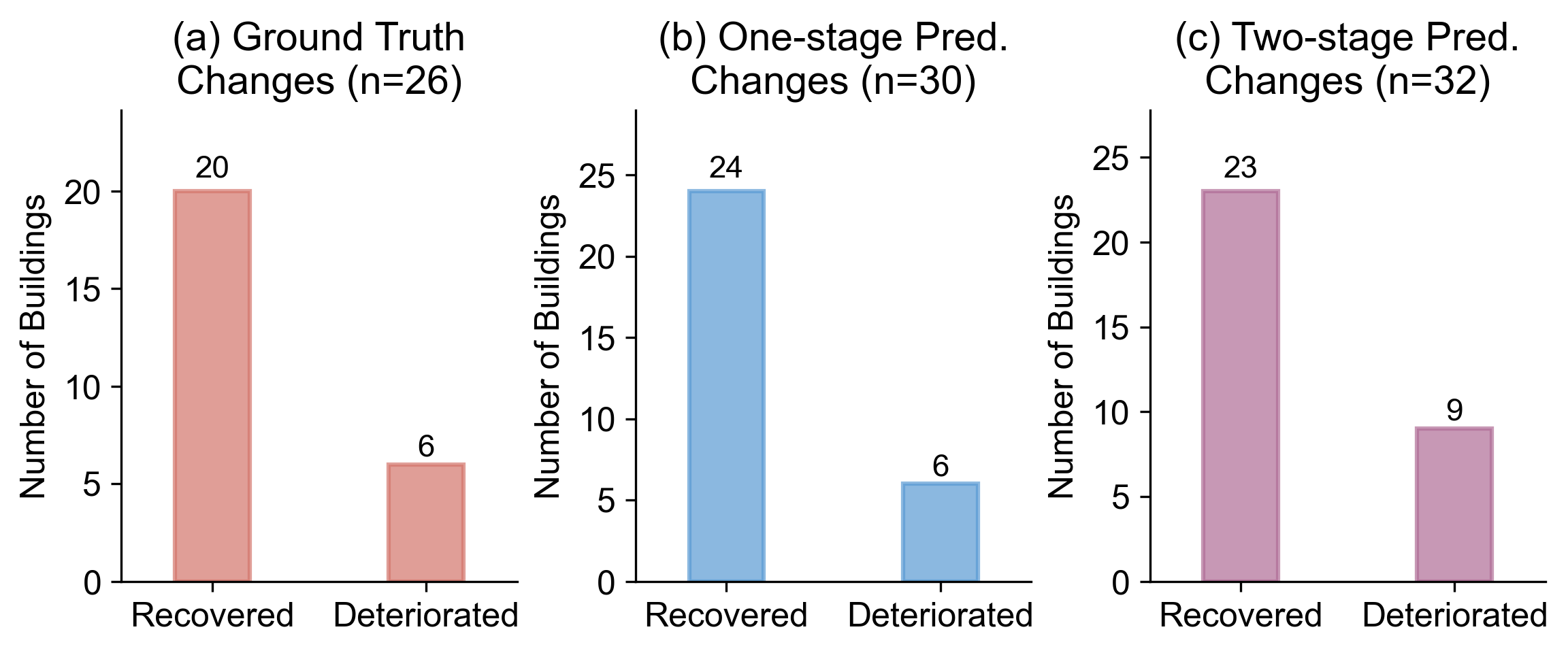} 
    \caption{Change analysis. (a) Ground-truth Recovered vs.\ Deteriorated counts (n=26). (b) One-stage predicted changes (n=30). (c) Two-stage predicted changes (n=32).}
    \label{fig:change-comparison}
\end{figure}

We analyze parcel-level occupancy change between V1 and V2 using the four classes defined in Table~\ref{tab:transitions} (Stable Occupied, Stable Not Occupied, Recovered, Deteriorated). Figure~\ref{fig:change-comparison} (a)--(c) show Recovered / Deteriorated counts: ground truth 20 / 6 (net +14), Two-stage 23 / 9 (net +14; matches net recovery), One-stage 24 / 6 (net +18; overstates recovery by +4). Thus, both strategies capture the direction of net improvement; the One-stage inflates recoveries, while the Two-stage balances additional true recoveries with extra deteriorations.

Figure~\ref{fig:change-comparison} (d) partitions parcel outcomes: Perfect Agreement: No Change (197; 82.1\%), Perfect Agreement: Change (17; 7.1\%), partial agreement where only one model matches ground truth (Two-stage only 8; 3.3\%, One-stage only 4; 1.7\%), and both models agree but wrong (14; 5.8\%). The last two groups can be a guide for the targeted review.

\begin{table}[!htbp]
\centering
\caption{Parcel-level occupancy change agreement categories (confident parcels with both visits).}
\label{tab:change-agreement}
\normalsize
\begin{tabular}{lcr}
\toprule
\textbf{Category} & \textbf{Count} & \textbf{Rate (\%)} \\
\midrule
Perfect Agreement: No Change & 197 & 82.1 \\
Perfect Agreement: Change & 17 & 7.1 \\
GT and Two-stage Agree & 8 & 3.3 \\
GT and One-stage Agree & 4 & 1.7 \\
Methods Agree, GT Differs & 14 & 5.8 \\
\midrule
Total & 240 & 100.0 \\
\bottomrule
\end{tabular}
\end{table}

\begin{figure}[!htbp]
    \centering
    \includegraphics[width=\textwidth]{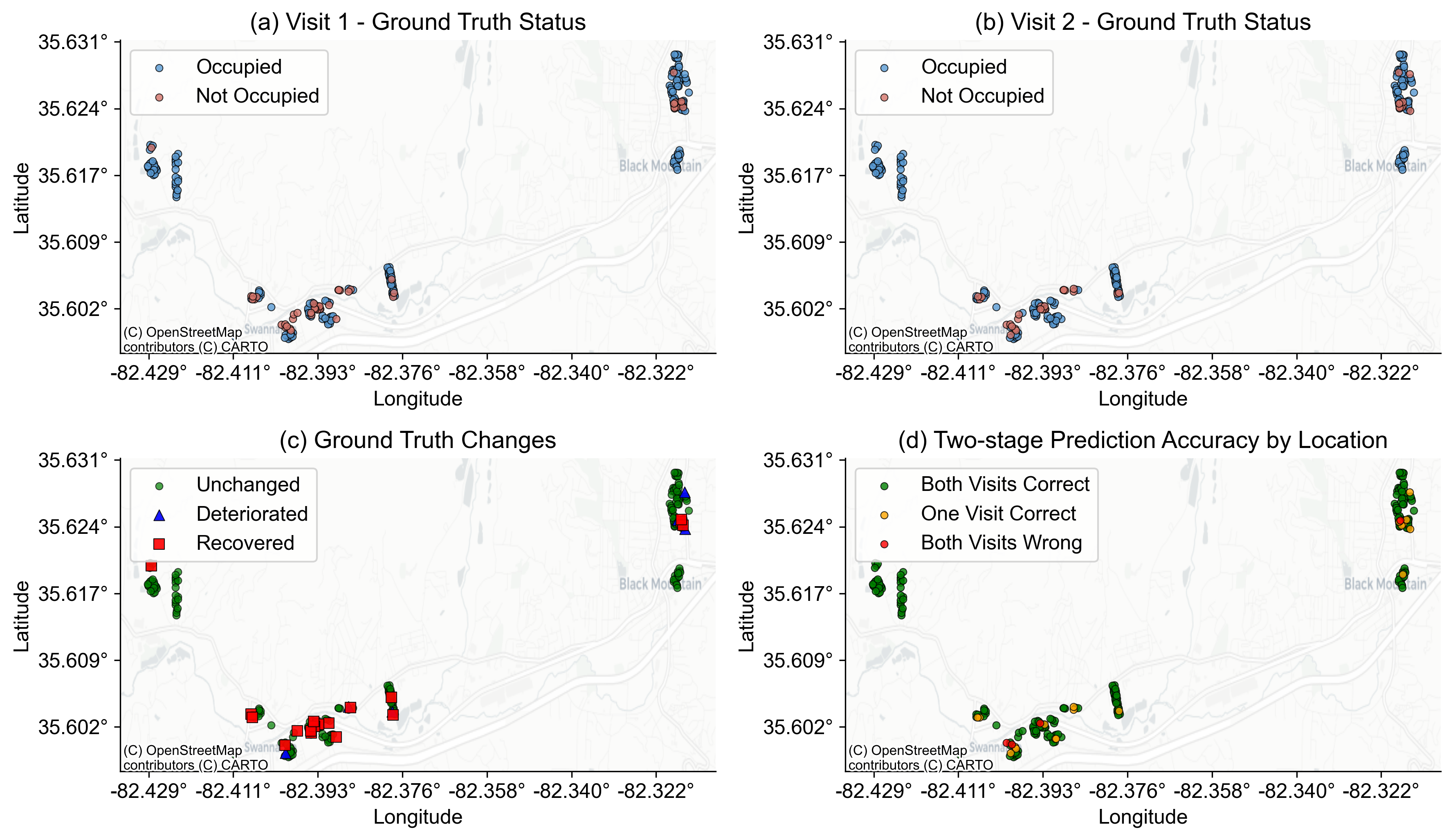}
    \caption{Geographic distribution of occupancy and model accuracy: (a) Visit 1 ground-truth status; (b) Visit 2 ground-truth status; (c) ground-truth change classes (Recovered, Deteriorated, Unchanged); (d) Two-stage prediction accuracy by parcel (both visits correct, one visit correct, both wrong).}
    \label{fig:recovery-hotspots}
\end{figure}

Spatial heterogeneity and remaining uncertainty are visualized in Figure~\ref{fig:recovery-hotspots}, which maps (a) Visit 1 ground-truth occupancy, (b) Visit 2 ground-truth occupancy, (c) ground-truth change classes (Recovered, Deteriorated, Unchanged), and (d) spatial distribution of Two-stage prediction accuracy (both visits correct, one visit correct, both wrong). This multi-panel view highlights localized pockets of recovery and deterioration and reveals where residual model errors cluster geographically, indicating priority areas for data or prompt refinement.

Together, these analyses contextualize raw change counts with spatial concentration, statistical confidence, and directional bias diagnostics, informing both methodological refinement and operational decision making.

We assessed spatial autocorrelation with the global Moran's $I$ for four spatial fields: parcel occupancy at Visit~1 and Visit~2, the ground-truth change field, and a binary accuracy indicator (1\,{=}\,correct, 0\,{=}\,incorrect). Positive $I$ with small $p$-values indicates spatial clustering rather than spatial randomness. Table~\ref{tab:moran} summarizes the results.

\begin{table}[!htbp]
\centering
\caption{Global Moran's $I$ statistics for parcel-level layers. Positive values indicate spatial clustering. 
The expected value under the randomization null is $E[I] = -1/(n-1)$; here $E[I]\approx -0.0042$ because $n=240$.}

\label{tab:moran}
\normalsize
\begin{tabular}{lccc}
\toprule
\textbf{Layer} & \textbf{Moran's $I$} & \textbf{Expected $I$} & \textbf{$p$-value} \\
\midrule
Visit 1 occupancy status   & 0.3704 & $-0.0042$ & 0.0000 \\
Visit 2 occupancy status   & 0.2834 & $-0.0042$ & 0.0000 \\
Ground truth changes       & 0.0799 & $-0.0042$ & 0.0048 \\
Prediction accuracy        & 0.0600 & $-0.0042$ & 0.0313 \\
\bottomrule
\end{tabular}
\end{table}

Under the standard randomization null, the expected global Moran's $I$ equals $-1/(n-1)$, where $n$ is the number of parcels in the analysis; using our common parcel mask ($n=240$) yields $E[I]\approx -0.0042$ for each layer. While the associated $p$-values confirm that clustering departs from spatial randomness, we interpret $I$ primarily as an effect size and rely on local statistics and maps to indicate where clustering concentrates.

These statistics indicate significant clustering for all four layers. Clustering is strongest for occupancy states (Visit~1 $I{=}0.3704$, Visit~2 $I{=}0.2834$) and weaker, but still significant, for the change field ($I{=}0.0799$) and the accuracy surface ($I{=}0.0600$). Practically, residual errors are not uniformly scattered; they concentrate in localized pockets, which supports prioritizing targeted audits of those neighborhoods.

\section{Discussion}\label{sec:discussion}

Our results suggest that separating perception (attribute extraction) from decision-making (conservative reasoning) is a pragmatic design choice for post-disaster occupancy inference. While the two-stage strategy yields higher recall and slightly higher F1/Cohen's $\kappa$ at the selected operating point, paired tests do not detect a statistically significant difference compared to the one-stage baseline. We therefore interpret the primary advantage of the two-stage design not as raw performance, but as controllability and auditability. By exposing intermediate attributes, this approach allows operators to diagnose failure modes, tune decision policies (e.g., risk thresholds), and prioritize targeted human review in spatially clustered error pockets. 

Beyond model architecture, the pipeline offers significant operational leverage. The separation of perception and reasoning mirrors the practical need to scale evidence collection: 360° video capture allowed for the rapid assessment of 914 residential parcels with limited field time. However, the performance drop in Visit 2 highlights a persistent challenge: post-repair scenes exhibit subtler cues that increase ambiguity. In such regimes, a conservative reasoning stage helps dampen the impact of uncertain visual evidence, though performance ultimately remains bounded by the quality of the underlying imagery.

\section{Conclusion}
\label{sec:conclusion}
Automated occupancy monitoring is crucial for tracking community recovery after disasters. We presented a street-level pipeline that pairs panoramic rectification with vision-language prompting to assess occupancy and track changes at the parcel scale. By fusing 5.6K rectified video frames with a VLM-based reasoning engine, we demonstrated a workflow that produces decision-ready, auditable outputs while successfully reproducing ground-truth net recovery trends.

We compared a transparent one-stage rule against a two-stage approach that separates perception from reasoning. While both strategies achieved high agreement with ground truth, the two-stage design offers superior interpretability. It allows for the isolation of visual perception errors from decision logic, facilitating the "tuning" of conservatism required in sensitive post-disaster contexts. Crucially, our spatial analysis revealed that residual errors are not random but spatially clustered; this finding validates the use of this pipeline to direct human QA/QC efforts to specific neighborhoods rather than diffuse random audits.

Limitations remain regarding geographic generalization and environmental variability, motivating future work in multi-region validation and temporal transition modeling. Ultimately, this framework provides a scalable alternative to door-to-door inspections, shifting field effort toward rapid right-of-way capture and providing recovery managers with an auditable, spatially-aware shortlist of recovery progress.

\section*{Acknowledgments}

\section*{Funding}
This material is supported by the National Science Foundation under the CAREER grant (No. 1846069) and the North Carolina General Assembly through the North Carolina Policy Collaboratory.

\section*{Data and Code availability}
The data that support the findings of this study are publicly available at \href{https://huggingface.co/datasets/Ymx1025/FacadeTrack}{Huggingface}, and the code is available at \href{https://github.com/YimingXiao98/RecovVision}{GitHub}.

\newpage
\appendix
\twocolumn
\section{Supplementary Information}\label{sec:appendix}

\subsection{Prompts for VLM-only strategy}

\begin{lstlisting}[style=prompt,caption={Actual vision-only prompt used (attribute extraction)},label={lst:vision-only-prompt}]
Analyze the image and answer with a JSON object using EXACTLY these keys and boolean values (true/false). Do not add, remove, or rename keys. Return only the JSON object, no prose.

{
  "house_destruction": true/false,
  "structural_damage": true/false,
  "exterior_debris": true/false,
  "open_doors_windows": true/false,
  "site_accessible": true/false,
  "exterior_mud": true/false,
  "emergency_markings": true/false,
  "major_repairs": true/false,
  "vehicle_presence": true/false
}
\end{lstlisting}

\lstdefinelanguage{json}{
  morestring=[b]",
  morecomment=[l]{//},
  morecomment=[s]{/*}{*/},
}

\begin{lstlisting}[style=jsonprompt,caption={Expected JSON string format (canonical schema)},label={lst:vision-only-json}]
{"house_destruction": false,
 "structural_damage": true,
 "exterior_debris": true,
 "open_doors_windows": false,
 "site_accessible": true,
 "exterior_mud": false,
 "emergency_markings": false,
 "major_repairs": false,
 "vehicle_presence": true}
\end{lstlisting}

\subsection{Prompts for decision stage (few-shot)}

\balance

\begin{lstlisting}[style=prompt,caption={Decision-stage prompt with few-shot examples (using canonical schema)},label={lst:decision-fewshot}]
You are an expert in post-disaster building occupancy assessment. Given a building's attributes in JSON (keys listed below), decide if it is 'Occupied' or 'Not Occupied'.

Consider all evidence: for example, some exterior mud may coexist with occupancy; parked vehicles can indicate occupancy when other signs are mixed; extensive roof or wall repairs may indicate temporary non-occupancy. If the evidence is mixed or unclear, prefer 'Not Occupied'. Be conservative: if there is significant indication that a building might not be occupied (e.g., destruction, inaccessibility, visible abandonment, or multiple risk factors), classify it as 'Not Occupied'. Classify as 'Occupied' only if the building appears livable and there are no clear signs of uninhabitability.

The JSON fields you will receive:
{
  "house_destruction": bool,
  "structural_damage": bool,
  "exterior_debris": bool,
  "open_doors_windows": bool,
  "site_accessible": bool,
  "exterior_mud": bool,
  "emergency_markings": bool,
  "major_repairs": bool,
  "vehicle_presence": bool
}

Here are some examples:

Example 1:
{
  "house_destruction": false,
  "structural_damage": false,
  "exterior_debris": false,
  "open_doors_windows": false,
  "site_accessible": true,
  "exterior_mud": false,
  "emergency_markings": false,
  "major_repairs": false,
  "vehicle_presence": true
}
Occupied

Example 2:
{
  "house_destruction": true,
  "structural_damage": true,
  "exterior_debris": true,
  "open_doors_windows": false,
  "site_accessible": true,
  "exterior_mud": false,
  "emergency_markings": false,
  "major_repairs": false,
  "vehicle_presence": false
}
Not Occupied

Example 3:
{
  "house_destruction": false,
  "structural_damage": true,
  "exterior_debris": true,
  "open_doors_windows": true,
  "site_accessible": true,
  "exterior_mud": false,
  "emergency_markings": false,
  "major_repairs": false,
  "vehicle_presence": true
}
Not Occupied

Example 4:
{
  "house_destruction": false,
  "structural_damage": true,
  "exterior_debris": false,
  "open_doors_windows": false,
  "site_accessible": true,
  "exterior_mud": false,
  "emergency_markings": false,
  "major_repairs": true,
  "vehicle_presence": false
}
Not Occupied

Example 5:
{
  "house_destruction": false,
  "structural_damage": false,
  "exterior_debris": false,
  "open_doors_windows": false,
  "site_accessible": true,
  "exterior_mud": false,
  "emergency_markings": true,
  "major_repairs": false,
  "vehicle_presence": false
}
Not Occupied

Example 6:
{
  "house_destruction": false,
  "structural_damage": false,
  "exterior_debris": false,
  "open_doors_windows": false,
  "site_accessible": true,
  "exterior_mud": false,
  "emergency_markings": false,
  "major_repairs": false,
  "vehicle_presence": false
}
Occupied

Example 7:
{
  "house_destruction": true,
  "structural_damage": false,
  "exterior_debris": false,
  "open_doors_windows": false,
  "site_accessible": true,
  "exterior_mud": false,
  "emergency_markings": false,
  "major_repairs": false,
  "vehicle_presence": false
}
Not Occupied

Example 8:
{
  "house_destruction": false,
  "structural_damage": false,
  "exterior_debris": false,
  "open_doors_windows": false,
  "site_accessible": false,
  "exterior_mud": false,
  "emergency_markings": false,
  "major_repairs": false,
  "vehicle_presence": false
}
Not Occupied

Example 9:
{
  "house_destruction": false,
  "structural_damage": false,
  "exterior_debris": false,
  "open_doors_windows": false,
  "site_accessible": true,
  "exterior_mud": true,
  "emergency_markings": false,
  "major_repairs": false,
  "vehicle_presence": true
}
Occupied

Now, decide for this building:
{...JSON from vision model...}

Output only one token: 'Occupied' or 'Not Occupied'.
\end{lstlisting}

\onecolumn


\clearpage
\begingroup
  \sloppy             
  \bibliography{references}
\endgroup

\end{document}